\documentclass[11pt,a4paper]{article}
\usepackage[hyperref]{acl2021}
\usepackage{times}
\usepackage{latexsym}

\usepackage{microtype}

\usepackage{bm}
\usepackage{subfigure}
\usepackage{amssymb}
\usepackage{amsmath}
\usepackage{arydshln}
\usepackage{multirow}
\usepackage{graphicx}

\usepackage{lipsum}

\aclfinalcopy % Uncomment this line for the final submission
\def\aclpaperid{3860} %  Enter the acl Paper ID here

\title{Guiding Teacher Forcing with Seer Forcing \\ for Neural Machine Translation}

\author{Yang Feng\textsuperscript{\rm 1,2}~~~Shuhao Gu\textsuperscript{\rm 1,2}~~~Dengji Guo\textsuperscript{\rm 1,2}~~~Zhengxin Yang\textsuperscript{\rm 1,2}~~~Chenze Shao\textsuperscript{\rm 1,2} \\ 
{\textsuperscript{\rm 1} {Key Laboratory of Intelligent Information Processing}} \\ Institute of Computing Technology, Chinese Academy of Sciences (ICT/CAS) \\
{ \textsuperscript{\rm 2} {University of Chinese Academy of Sciences, Beijing, China}} \\
\{\texttt{fengyang, gushuhao19b, guodengji19s}\}@ ict.ac.cn\\ \{\texttt{yangzhengxin17z, shaochenze18z}\}@ ict.ac.cn
}

\date{}
\begin{document}
\maketitle

\newcommand\blfootnote[1]{
\begingroup
\renewcommand\thefootnote{}\footnote{#1}%
\addtocounter{footnote}{-1}%
\endgroup
}

\blfootnote{The code: https://github.com/ictnlp/SeerForcingNMT} 

\begin{abstract}
Although teacher forcing has become the main training paradigm for neural machine translation, it usually makes predictions only conditioned on past information, and hence lacks global planning for the future. To address this problem, we introduce another decoder, called seer decoder, into the encoder-decoder framework during training, which involves future information in target predictions. Meanwhile, we force the conventional decoder to simulate the behaviors of the seer decoder via knowledge distillation. In this way, at test the conventional decoder can perform like the seer decoder without the attendance of it. Experiment results on the Chinese-English, English-German and English-Romanian translation tasks show our method can outperform competitive baselines significantly and achieves greater improvements on the bigger data sets. Besides, the experiments also prove knowledge distillation the best way to transfer knowledge from the seer decoder to the conventional decoder compared to adversarial learning and L2 regularization. 
\end{abstract}

\section{Introduction}

Neural machine translation (NMT) \cite{kalchbrenner2013recurrent,sutskever2014sequence,bahdanau2014neural,gehring2017convolutional,vaswani2017attention} has achieved great success and is drawing larger attention recently. Most NMT models are under the attention-based encoder-decoder framework which assumes
 there is a common semantic space between the source and target languages. The encoder encodes the source sentence to the common space to get its meaning,  and the decoder projects the source meaning to the target space to generate corresponding target words. Whenever generating a target word at a time step, the decoder needs to retrieve the attended source information and then decodes into a target word. The underline principle which makes sure the framework works is that the information hold by the source sentence and its target counterpart is equivalent. Thus the translation procedure can be considered to decompose source information into different pieces and then to convert each piece to a proper target word according to bilingual context. When all the information encoded in the source sentence is throughly processed, the whole translation has been generated.
 
Neural machine translation models are usually trained via maximum likelihood estimation (MLE) \cite{johansen1990maximum} and the operation form is known as {\em teacher forcing} \cite{williams1989learning}. The teacher forcing strategy performs one-step-ahead predictions with the past ground truth words fed as context and forces the distribution of the next prediction to approach a 0-1 distribution where the probability of the next ground truth word corresponds to 1 and others to 0. In this way, the predicted sequence is trained to be close to the ground truth sequence. 
From the perspective of information division, the function of teacher forcing is to teach the translation model how to segment source information and derive the ground truth word from the source information at a maximum probability.
 
However, teacher forcing can only provide up-to-now ground truth words for one-step-ahead predictions and hence lacks global planning for the future. This will result in local optimization especially when the next prediction is highly related to the future. Besides, as the translation grows, the previous prediction errors will be accumulated and affect later predictions \cite{zhang2019bridging}. This is the important reason why NMT models cannot always produce the ground truth sequence during training. Therefore, it is more possible to achieve global optimization by getting to know the future ground truth words. This can lead to better cross-attention to the source sentence and thus better information devision. But unfortunately, ground truth can be only obtained during training and we cannot inference with future ground truth at test.
 
 To address this problem, we introduce an additional seer decoder into the encoder-decoder framework to integrate future information.
 During training, the seer decoder is used to guide the behaviors of the conventional decoder while at test the translation model only inferences with the conventional decoder without introducing any extra parameters and calculation cost.
% we propose a method to maintain a seer decoder and a teacher decoder during training while only inference with the teacher decoder at test, so that no extra parameters and calculation cost  are introduced at test.
 Specifically, the conventional decoder  only gets past information participating in the next prediction, %It only has past information participating in the next prediction, which is past ground truth words during training while previously self-generated words at test. 
while the seer decoder has both the past and future ground truth words engaged in the next prediction. Both decoders are trained to generate ground truth via MLE and meanwhile the conventional decoder is forced to simulate the behaviors of the seer decoder via knowledge distillation \cite{bucilua2006model,hinton2015distilling}. In this way, at test the conventional decoder can perform like the seer decoder as if it knew the future translation.

 We conducted experiments on two small data sets (Chinese-English and English-Romanian) and two big data sets (Chinese-English and English-German) and the experiment results show that our method can outperform strong baselines on all the data sets. In addition, we also compared different mechanisms of transferring knowledge and found that knowledge distillation is more effective than adversarial learning and L2 regularization. To the best of our knowledge, this paper is the first to explore the effects of the three mechanisms simultaneously in machine translation.

\begin{figure}[!t]
    \centering
    \includegraphics[scale=0.36]{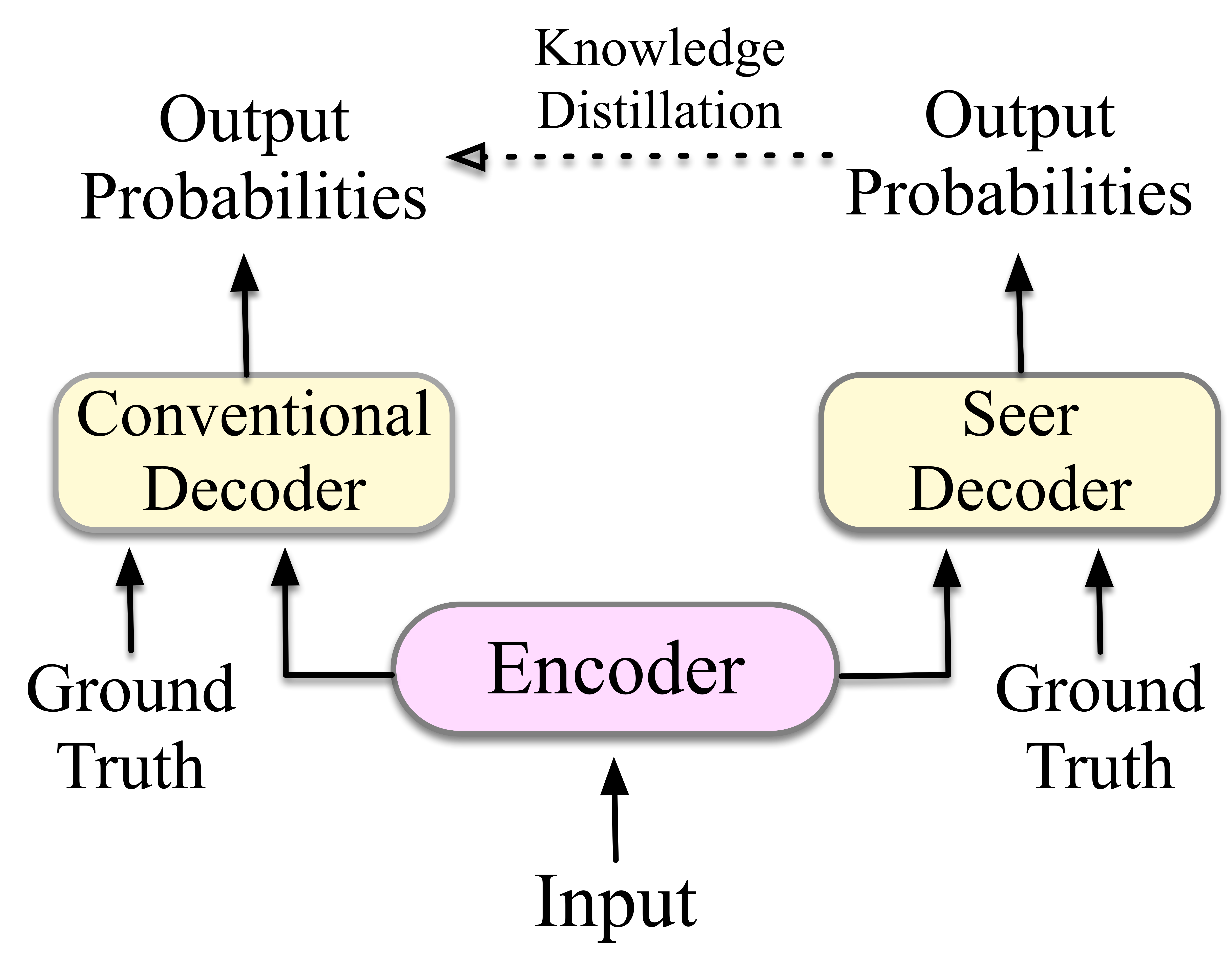}
    \caption{The architecture of the proposed method}
    \label{fig-model}
\end{figure}

\section{The Proposed Method}

%Our work aims to generate translation as if the decoder could foresee what target words would be produced in the next. The intuition is that if we integrate the future ground truth information into the self-attention representation for the next prediction, we can get better cross attention to the source sentence and hence better translation. But we cannot tell the seer decoder the current ground truth word, otherwise the translation model will not learn how to derive the true target word, but only learn to copy the right word from the target presentations.

We introduce our method on the basis of  {\em Transformer} which is under the encoder-decoder framework \cite{vaswani2017attention}. Our model consists of three components: the encoder, the conventional decoder and the seer decoder. 
The architecture is shown in Figure \ref{fig-model}. The encoder and the conventional decoder work in the same way as the corresponding components of Transformer do. The seer decoder integrates future ground truth information into its self-attention representation and calculates cross-attention over source hidden states with the self-attention representation as the query. During training, the encoder is shared by the two decoders and both decoders perform predictions to generate ground truth. 
The behaviors of the conventional decoder are guided by the seer decoder via knowledge distillation. If the conventional decoder can predict a similar distribution as the seer decoder, we think the conventional decoder performs like the seer decoder. Then we can only use the conventional decoder for test.

The details of the encoder and the conventional decoder can be got from \citet{vaswani2017attention}. Assume the input sequence is $\bm{\mathrm{x}}=(x_1 ,..., x_J)$,  the ground truth sequence is $\bm{\mathrm{y}}^*=(y^*_1 ,..., y^*_I)$ and the generated translation is $\bm{\mathrm{y}}=(y_1 ,..., y_I)$.
 We will give more description to the seer decoder and the training in what follows. 

\subsection{The Seer Decoder}

\begin{figure}[!t]
    \centering
    \includegraphics[scale=0.28]{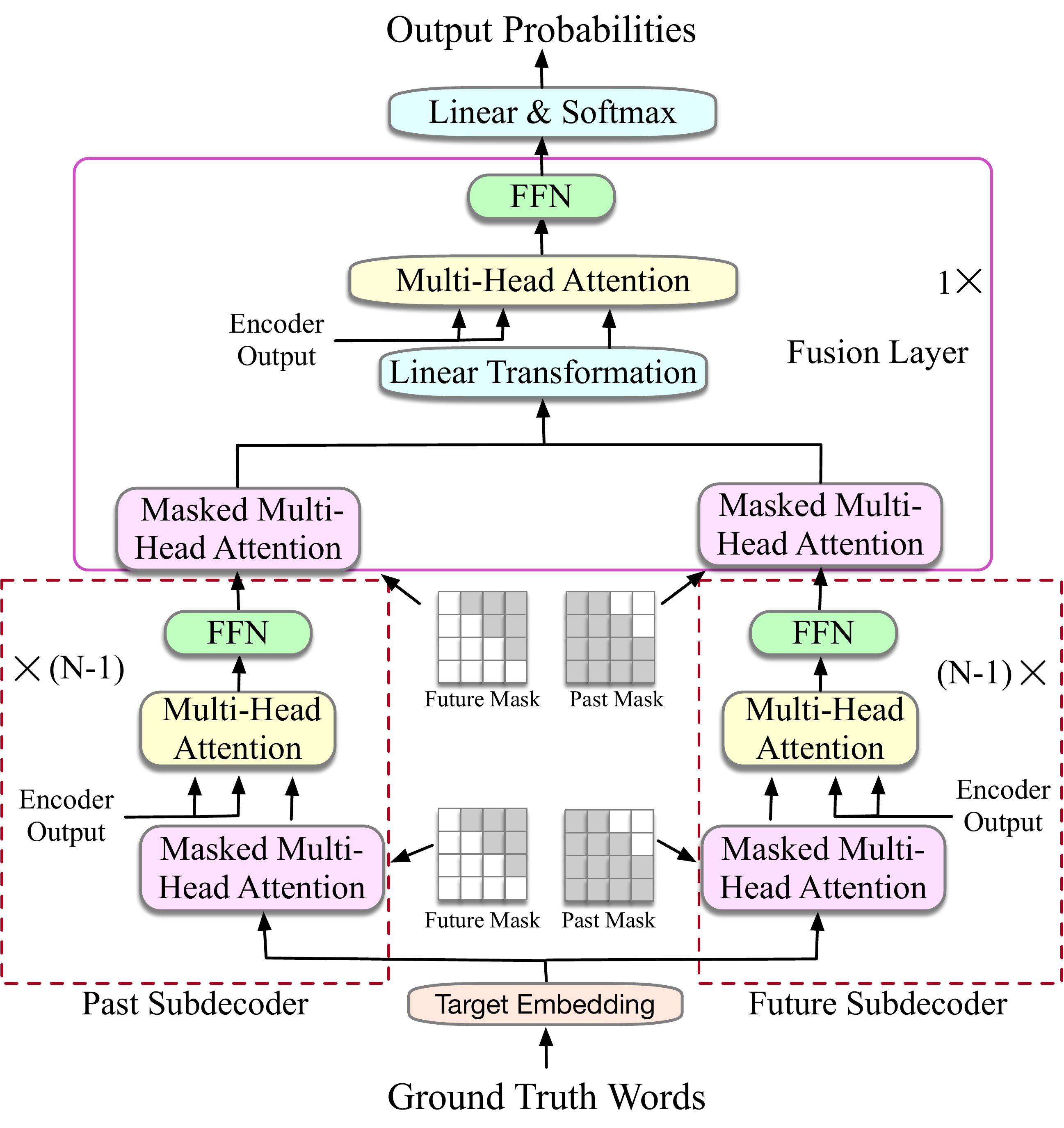}
    \caption{The architecture of the seer decoder}
    \label{fig-seerdecoder}
\end{figure}

Although we feed the future ground truth words to the seer decoder, we will not tell it the next ground truth word to be generated, in case it will only learn a copy operation, not how to derive a word. Considering efficiency, the seer decoder does not integrate the past and future ground truth information with a unique decoder , but two separate subdecoders. As a result,  the seer decoder consists of three components: the past subdecoder, the future subdecoder and the fusion layer. The architecture of the seer decoder is given in Figure \ref{fig-seerdecoder}. The past and future subdecoders are employed to decode the past and future ground truth information into hidden states respectively and the fusion layer is used to fuse the output of the past and future subdecoders and calculate the final hidden state for the next prediction. 

The past subdecoder is composed of  $N \mathcal{-} 1$ layers and each layer has three sublayers which are the multi-head sublayer, the cross-attention sublayer and the feed-forward network (FNN) sublayer, the same as Transformer. The multi-attention sublayer accepts the whole ground truth sequence as the input and applies  a mask matrix $\bm{\mathrm{M}}_p$ to make sure only the past ground truth words attend the self-attention. Specifically, to generate the $i$-th target word, its corresponding mask vector in the mask matrix $\bm{\mathrm{M}}_p$ is set to mask the words $y^*_i, y^*_{i+1}, ...,   y^*_I$. Then after the cross-attention sublayer and the FFN sublayer, the past subdecoder output a sequence of past hidden states, the packed matrix of which is denoted as $\bm{\mathrm{H}}_p$.

The future subdecoder has the same structure as the past subdecoder except for the mask matrix. The future subdecoder also has the whole ground truth sequence as the input but employs a different mask matrix $\bm{\mathrm{M}}_f$ to only remain the future ground truth information. To generate the $i$-th target word, the corresponding mask vector in $\bm{\mathrm{M}}_f$ masks the words $y^*_1, ..., y^*_{i-1}, y^*_i$. The packed matrix of the future hidden states generated by the future subdecoder is denoted as $\bm{\mathrm{H}}_f$.

The fusion layer is composed of four sublayers:  the multi-head sublayer, the linear sublayer, the cross-attention sublayer and the FFN sublayer. Except the linear sublayer, the rest three sublayers works in the same way as Transformer does. The multi-head sublayer encodes the outputs of the past and future subdecoders separately with the mask matrix $\bm{\mathrm{M}}_p$ and $\bm{\mathrm{M}}_f$, and the packed matrix of their output are denoted as $\bm{\mathrm{H}}'_p$ and $\bm{\mathrm{H}}'_f$ respectively. Then we reverse the order of the vectors in $\bm{\mathrm{H}}'_f$ to get $\bm{\mathrm{H}}''_f$, so that the same index in $\bm{\mathrm{H}}'_p$ and $\bm{\mathrm{H}}''_f$ can correspond to the past and future representation needed for the same prediction. Assume
$\bm{\mathrm{H}}'_f = [\bm{\mathrm{h}}'_{f1}; \bm{\mathrm{h}}'_{f2}; ...;  \bm{\mathrm{h}}'_{fI}]$, then its reversed matrix is 
$\bm{\mathrm{H}}''_f = [\bm{\mathrm{h}}'_{fI}; ...;  \bm{\mathrm{h}}'_{f2}; \bm{\mathrm{h}}'_{f1}] $.
The linear sublayer fuses $\bm{\mathrm{H}}'_p$ and $\bm{\mathrm{H}}''_f$ via a linear transformation as
\begin{equation}
\bm{\mathrm{A}} = \bm{\mathrm{W}}_p \bm{\mathrm{H}}'_p + \bm{\mathrm{W}}_f \bm{\mathrm{H}}''_f  \label{eq-fuse}
\end{equation}
Now we can think each representation in the matrix $\bm{\mathrm{A}}$ incorporates the past and future information for its corresponding prediction. Then after the cross-attention sublayer over the outputs of the encoder and then the FFN sublayer, we can get the target hidden states produced by the seer decoder as $\bm{\mathrm{S}}_s=[\bm{\mathrm{s}}_{s_1}; ...; \bm{\mathrm{s}}_{s_I}]^T$. Then the probability to generate the target word $y_i$ is
\begin{equation}
p_s(y_i | \bm{\mathrm{y}}^*_{>i}, \bm{\mathrm{y}}^*_{<i}, \bm{\mathrm{x}}) \propto \exp{(\bm{\mathrm{W}}_o \bm{\mathrm{s}}_{si})} \label{eq-seerout}
\end{equation}

Note that the past and the future subdecoders share the same set of parameters, and the same linear transformation matrix $\bm{\mathrm{W}}_o$ %(as in Equation \ref{eq-outprojection}) 
is applied to the outputs of the conventional and seer decoders.

\section{Training}

In our method, only the conventional decoder is employed for test and the seer decoder is only used to guide the conventional decoder during training.
Given a sentence pair $\langle \bm{\mathrm{x}},\bm{\mathrm{y}}^* \rangle$ in the training set, the conventional decoder and the seer decoder can predict a distribution for target position $i$ as $p_c(y_i | \bm{\mathrm{y}}^*_{<i}, \bm{\mathrm{x}}) $ and $p_s(y_i | \bm{\mathrm{y}}^*_{>i}, \bm{\mathrm{y}}^*_{<i}, \bm{\mathrm{x}}) $, respectively. The two decoders are both trained by comparing its predicted distribution with the 0-1 distribution of the ground truth word by minimizing the cross entropy, that is to maximize the likelihood of the corresponding ground truth word. As the two decoders involve different information for next prediction, we call the training strategy {\em teacher forcing} and {\em seer forcing}, respectively. 
The cross-entropy loss for the conventional decoder is 
\begin{equation}
\mathcal{L}_c = -\sum_{k=1}^\mathrm{K} \sum_{i=1}^{I_k} \log p_c(y^*_i | \bm{\mathrm{y}}^*_{<i}, \bm{\mathrm{x}}) ,
\end{equation}
and the cross-entropy loss for the seer decoder is
\begin{equation}
\mathcal{L}_s = -\sum_{k=1}^\mathrm{K} \sum_{i=1}^{I_k} \log p_s(y^*_i | \bm{\mathrm{y}}^*_{>i}, \bm{\mathrm{y}}^*_{<i}, \bm{\mathrm{x}}) .
\end{equation}
where $\mathrm{K}$ is the size of the training set and $I_k$ is the length of the $k$-th target sentence.

The conventional decoder is further trained to get close to the distribution of the seer decoder via knowledge distillation.
In knowledge distillation, the conventional decoder ({\em the student}) has to not only match the one-hot ground truth word, but fit the distribution over the target vocabulary V drawn by the seer decoder ({\em the teacher}). The knowledge distillation loss can be formalized as 
\begin{align}
 \mathcal{L}_{kd} =  -\sum_{k=1}^\mathrm{K} \sum_{i=1}^{I_k} \sum_{l=1}^{|\mathrm{V}|} 
  & \ p_s(y_i=l | \bm{\mathrm{y}}^*_{>i}, \bm{\mathrm{y}}^*_{<i}, \bm{\mathrm{x}}) \nonumber \\
  & \times \log p_c(y_i=l | \bm{\mathrm{y}}^*_{<i} , \bm{\mathrm{x}}) 
\end{align}
where $|\mathrm{V}|$ is the size of the target vocabulary.

The final training loss is
\begin{equation}
\mathcal{L} = \mathcal{L}_s + \lambda \mathcal{L}_c + (1-\lambda) \mathcal{L}_{kd} \ .  \label{eq-loss}
\end{equation}
Different from the conventional knowledge distillation which first trains the teacher via cross entropy against ground truth, then fixes the teacher and only trains the student, we train all the parameters from the scratch, but we still follow the above rule to keep the teacher (i.e. the seer decoder) unchanged in the process of distillation. To do this, we do not update the parameters of the seer decoder through the loss $\mathcal{L}_{kd}$, that is, we only back propagate gradients to the seer decoder through $\mathcal{L}_{s}$, but not through $\mathcal{L}_{kd}$.

\section{Related Work}

%Our work introduces a seer decoder which holds both the past and future ground truth to act as the mentor of the teacher decoder. In this way, the model can achieve a better optimization and inference as regularizational without extra parameters.

Reinforcement-learning-based methods also encode future information in the rewards to supervise fine-tuning of the translation model. The rewards are worked out either by sampling future translation with the REINFORCE algorithm \cite{williams1992simple,yu2017seqgan,yang2018improving,shao2019retrieving}, or by directly calculating a value with the actor-critic algorithm \cite{bahdanau2016actor,li2017learning}. This set of methods only give a weak supervision to the NMT model through rewards and suffer from unstable training. In contrast, \citet{shao2018greedy} propose to train autoregressive NMT with the probabilistic n-gram based GLEU \cite{wu2016google} and \citet{shao2020minimizing} propose to minimize the bag-of-ngrams difference for non-autoregressive NMT so that the two methods can abandon reinforcement learning and perform training directly by gradient descent. 

%Some researchers also have the whole ground truth sequence involved during training to make global planning by using a sequence level reward. \citet{yang2018improving} extend the work of \citet{yu2017seqgan} to the conditional case of machine translation. They introduce BLEU as an additional reward function to the SeqGAN model and train the model under the framework of reinforcement learning, specifically the REINFORCE algorithm \cite{williams1992simple}. \citet{shao2018greedy} cut off  the discriminator of SeqGAN and train the model with the probabilistic n-gram based GLEU \cite{wu2016google}, so that they can abandon reinforcement learning and train their model directly by gradient descent. This set of work needs the help of a sequence level reward.

Another set of methods introduce future information into inference with additional pass of decoding or extra components at test.
\citet{niehues2016pre}, \citet{xia2017deliberation}, \citet{hassan2018achieving} and \citet{zhang2018asynchronous} proposed a two-pass decoding algorithm to first generate a draft translation and then generate final translation referring to the draft.
 \citet{geng2018adaptive} expand this line of methods by performing an adaptive multi-pass decoding where the number of decoding passes is determined by a policy network.
 \citet{liu2016agreementa}, \citet{liu2016agreementb},  \citet{hoang2017towards}, \citet{zhang2019regularizing} and \citet{he2019multi} perform bidirectional decoding simultaneously and the two decoders correlate to each other via an agreement term or a regularization term in the loss.
\citet{zhou2019synchronous} , \citet{zhou2019sequence} and \citet{zhang2019synchronous} also maintain a forward decoder and a backward decoder to decode simultaneously but they interact to each other when making predictions. \citet{zhang2019future} introduce a future-aware vector at test which is learned via the knowledge distillation framework during training.
The difference between this set of methods and our method is that our method does not require any other cost at test and is easy to use.
 
%%\citet{zheng2018modeling} and \citet{zheng2019dynamic} work in another perspective by modeling past and future information for the source to help the decoder focus on untranslated source information.

There are some other works which integrate future information during training while only perform one-pass decoding. \citet{serdyuk2018twin} introduce a twin network to perform bidirectional decoding simultaneously during training and force the hidden states generated by the two decoders to be consistent, then at inference it can only use the forward decoder. But in this method the two decoders act as a counterpart to each other and no decoder plays a role of teacher, which determines that it can only be trained via  L$_2$ regularization, not knowledge distillation which has proven in the experiments more effective than L$_2$ regularization. 
%, aiming that the forward decoder can remain the information to generate the future translation.
\citet{feng2020modeling} introduce an evaluation module to give each translation more reasonable evaluation when it cannot match the ground truth. The evaluation is conducted from the perspective of fluency and faithfulness which both need the participation of past and future information. The difference from the method proposed in this paper is their method uses self-generated translation as past information and does not train with knowledge distillation.

Some researchers work in another perspective by introducing future information. \citet{zhang2020future} propose to
 employ future source information to guide simultaneous machine translation with knowledge distillation, so that the incompleteness of source can be mitigated. \citet{zheng2018modeling} and \citet{zheng2019dynamic}  propose to model past and future information for the source to help the decoder focus on untranslated source information. 
 
%There are some researchers that have also found the importance of future information in other tasks, e.g., \citet{qi2020prophetnet} present a new sequence-to-sequence pre-training model which is optimized with $n$-step ahead prediction participating in, and \citet{zhang2020improving} introduce a guiding network into text generation to model the future generation environment, which can return rewards to guide next token generation.

%%%%%
%Some researchers working at the global optimization by introducing another decoder to provide the translation of untranslated part. \citet{zhang2018asynchronous} and \citet{hassan2018achieving} proposed a two-pass decoding algorithm which first performs the first-pass decoding to get a draft and then conducted the second-pass decoding to get refined translation based on the draft. \citet{zhang2019synchronous} propose a synchronous bidirectional decoding algorithm which perform left-to-right decoding and right-to-left decoding simultaneously so that the two decoders can communicate each other. All these methods requires decoding twice to generate two translations in two-pass or interactively.

%Some other researchers also apply adversarial learning to NMT. \citet{cheng2018towards} employ a discriminator to distinguish the translation with and without noise so that the decoder with noise input can acts like the clean decoder with the help of adversarial leaning.

\section{Experiments}

%In the experiment section, we will first verify the rationality of adopting the seer forcing and the discriminator, and further make some analysis to our method, and finally compare our method with some baselines.

\subsection{Settings}

\subsubsection{Data Preparation}

We conducted experiments on two small data sets and two big data sets. 

\textbf{Small Data Sets} 

\textbf{Chinese$\rightarrow$English} The training set consists of about 1.25M sentence pairs from LDC corpora with 27.9M Chinese words and 34.5M English words respectively\footnote{The corpora include LDC2002E18, LDC2003E07, LDC2003E14, Hansards portion of LDC2004T07, LDC2004T08 and LDC2005T06.}. We used MT02 for validation and MT03, MT04, MT05, MT06, MT08 for test. We tokenized and lowercased English sentences using the Moses scripts\footnote{http://www.statmt.org/moses/}, and segmented the Chinese sentences with the Stanford Segmentor\footnote{https://nlp.stanford.edu/}. The two sides were further segmented into subword units using Byte-Pair Encoding(BPE)~\cite{sennrich2016neural} with 30K merge operations. 32K size of the Chinese dictionary and 29K size of the English dictionary were built for the two sides. 

\textbf{English$\rightarrow$Romanian} We used the preprocessed version of WMT16 En-Ro dataset released by \citet{lee2018deterministic} which includes 0.6M sentence pairs. We used news-dev 2016 for validation and news-test 2016 for test. The two languages share the 35K size of the joint vocabulary generated with 40K merge operations of BPE on the combined data.

\textbf{Big Data Sets}

\textbf{Chinese$\rightarrow$English}  The training data is from WMT 2017 Zh-En translation tasks that contains 20.18M sentence pairs after deleting duplicate ones. The newsdev2017 was used as the development set and newstest2017 was used as the test set. To avoid the effects of the translationese~\cite{abs-1906-09833}, we also tested the methods on the newstest2019 test set. We tokenized and truecased the English sentences with Moses scripts. For the Chinese data, we performed word segmentation by using Stanford Segmenter. 32K BPE sizes were applied to the training data seperately and then we filtered out the sentences which are longer than 128 sub-words. 44K size of the Chinese dictionary and 33K size of the English dictionary were built based on the corresponding data.

\textbf{English$\rightarrow$German} The training data is from WMT2016 which consists of about 4.5M sentences pairs with 118M English words and 111M German words. The newstest2014 was used as the development set and newstest2016 and newstest2019 were used as the test sets.  The two languages share the 32K size of the joint vocabulary generated with 30K merge operations of BPE on the combined data.

\subsubsection{Systems} 

%We explored both the RNN-based and self-attention-based frame in our experiments.

%{\bf \textsc{RnnSearch}} Our in-house implementation of \citet{bahdanau2014neural} which is further enhanced with multi-head additive attention with 8 heads. The encoder is composed of 4 layers of bidirectional RNNs and the decoder is composed 2 layers of bidirectional RNNs. The word embedding dimension and the hidden size are both $512$. The parameters are initialized uniformly between [-0.08, 0.08] and updated by the Adam optimizer \cite{kingma2014adam} ($\beta_1=0.9$, $\beta_2=0.999$ and $\epsilon=1e^{-6}$) with 8000 warming-up steps and gradient regularization as $5.0$. Dropout is applied on embeddings, the output of the encode and the decoder at the rate of $0.5, 0.5, 0.5$ respectively. The training was conducted on $2$ GPUs and each batch is limited to 4096 tokens per GPU. At test beam size was set to 10.
\ 

{\bf \textsc{Transformer}} We used an open-source toolkit called {\em Fairseq-py} released by Facebook \cite{ott2019fairseq} which was implemented strictly following \citet{vaswani2017attention}. 

{\bf \textsc{RL-NMT}} We trained Transformer under the reinforcement learning framework using the REINFORCE algorithm \cite{williams1992simple} with the BLEU as the rewards. The implementation details for the RL part is the same as \citet{yang2018improving}.

{\bf \textsc{ABDNMT}} Our implementation of \citet{zhang2018asynchronous} based on Transformer.

{\bf \textsc{Twinnet}} Our implementation of \citet{serdyuk2018twin} based on Transformer. The weight of L$_2$ loss was $0.2$ \ .

{\bf \textsc{EVANMT}} Our implementation of \citet{feng2020modeling}.

{\bf \textsc{Seer+L$_2$}} Seer forcing with L$_2$ regularization. Similar to  \textsc{Twinnet}, we set $\mathcal{L}_2=\sum_{k=1}^\mathrm{K} \sum_{i=1}^{I_k}\|g(\bm{\mathrm{s}}_{ti}) - \bm{\mathrm{s}}_{si})\|_2$ where $g$ is a linear transformation. We first pretrained the two decoders together only with $\mathcal{L} = \mathcal{L}_t + \mathcal{L}_s$, then trained them with the loss of $\mathcal{L} = \mathcal{L}_t + \mathcal{L}_s + \alpha \mathcal{L}_2$  where $\alpha=0.2$, too. Please note that the L$_2$ loss did not update the seer decoder and the encoder so that the conventional decoder would approach the seer decoder, which followed \citet{serdyuk2018twin}.

{\bf \textsc{Seer+AL}} Seer forcing with adversarial learning. A discriminator is employed to distinguish the hidden state sequences generated by the conventional decoder and the seer decoder. The discriminator is based on CNN, implemented according to \citet{gu2019improving}. The translation model and the discriminator are trained jointly via a gradient reversal layer just like our method. The loss is $\mathcal{L} = \mathcal{L}_t + \mathcal{L}_s + \alpha \mathcal{L}_d$ where $\mathcal{L}_d$ is the loss of the discriminator and $\alpha=0.3$ on the EN$\rightarrow$RO data set and $\alpha=0.2$ on the other data sets.
%{\bf Our Method} Implemented based on Fairseq-py. For the weight $\lambda$ in Equation \ref{eq-loss},

\begin{table*}[ht!]
\centering
\resizebox{2.1\columnwidth}!{
\begin{tabular}{l   |   l l l l l l l l    |   l l l}
\hline
& \multicolumn{8}{|c|}{ \bf{CN$\rightarrow$EN} } &\multicolumn{3}{|c}{\bf{EN$\rightarrow$RO}} \\ 
& \bf \textbf{MT03} & \bf \textsc{MT04} & \bf \textsc{MT05} &  \bf \textsc{MT06}  & \bf \textsc{MT08} & \bf \textsc{AVG}  & \bf $\Delta$ & \bf \textsc{Time} &  \bf \textsc{WMT16} & \bf $\Delta$ & \bf \textsc{Time} \\
\hline
\hline
\bf \textsc{Transformer}  & 46.54 & 46.95 & 46.39 & 45.39 & 36.75 & 44.40 &  & 1.0 & 32.60 &  & 1.0  \\
\bf \textsc{RL-NMT}  & 45.75 & 47.41 & 46.44 & 47.08 & 37.65 & 44.87 &  {\em \small{+0.47}}  & 1.70 &  32.79 &  {\em \small{+0.19}} & 2.38 \\
\bf \textsc{ABDNMT}  & 47.16 & 47.58 & 46.77 & 45.97 & 36.43 & 44.78 &  {\em \small{+0.38}}  & 2.78 &  33.80 &  {\em \small{+1.20}} & 3.36 \\
\bf \textsc{Twinnet}  & 47.78 & 48.74 & \bf{48.59} & 46.65 & \bf{38.80} & 46.11 &  {\em \small{+1.71}}  & 2.56 &  33.79 & {\em \small{+1.19}} & 2.62 \\
\bf \textsc{EVANMT}  & 47.05 & 47.76 & 46.59 & 46.58 & 37.39 & 45.07 &  {\em \small{+0.67}}  & 2.19 &  33.29 & {\em \small{+0.69}} & 2.91 \\
\bf \textsc{Seer+L$_2$} & 47.98** & 48.66** & 48.16** & 47.02** & 38.64** & 46.09 & {\em \small{+1.69}} & 1.90 &  33.55** & {\em \small{+0.95}} & 1.83 \\
\bf \textsc{Seer+AL} & 47.91** & 48.38** & 47.97** & 47.04** & 38.18** & 45.89 & {\em \small{+1.49}} & 2.64 &  33.59** & {\em \small{+1.04}} & 2.35 \\
\bf {Our Method} & \bf{48.12**} & \bf{48.85**} & 48.25** & \bf{47.25**} & 38.71** & \bf{46.24} &  \bf{\em \small{+1.84}} &  1.92 & \bf{33.86**} &  \bf{\em \small{+1.26}} & 1.86 \\
% MT02: 31.41 52.53
\hline
\end{tabular}
}
\caption{BLEU scores on small data sets. ** mean the improvements over  \textsc{Transformer} is statistically significant \cite{collins2005clause} ($\rho < 0.01$, respectively).} \label{tab-total-small}
\end{table*}

\begin{table*}[ht!]
\centering
\resizebox{1.9\columnwidth}!{
\begin{tabular}{l   |   l l l l l    |   l l l l l}
\hline
& \multicolumn{5}{|c|}{ \bf{CN$\rightarrow$EN} } &\multicolumn{5}{|c}{\bf{EN$\rightarrow$DE}} \\ 
& \bf \textsc{2017} & \bf $\Delta$ & \bf \textsc{2019} & \bf $\Delta$ & \bf \textsc{Time} &  \bf \textsc{2016} & \bf $\Delta$ &  \bf \textsc{2019} & \bf $\Delta$ & \bf \textsc{Time} \\
\hline
\hline
\bf \textsc{Transformer} & 23.75 & & 26.00 & & 1.0 & 33.49 && 36.20 && 1.0 \\
\bf \textsc{Twinnet} & 23.39 & \em \small{-0.36} & 26.09 & \em \small{+0.09} & 2.58 & 33.05 & \em \small{-0.44} & 35.69 & \em \small{-0.51} & 2.57  \\

\bf \textsc{EVANMT} & 23.80 & \em \small{+0.05} & 26.07 & \em \small{+0.07} & 2.37 & 34.00 & \em \small{+0.51} & 37.25 & \em \small{+1.05} & 2.48  \\

\bf \textsc{Seer+L$_2$} & 23.95 & \em \small{+0.20} & 25.82 & \em \small{-0.18} &1.93 & 33.58 & \em \small{+0.09} & 36.65 & \em \small{+0.45} & 1.53 \\
\bf \textsc{Seer+AL} & 24.01 & \em \small{+0.26} & 26.47* & \em \small{+0.47} &2.29 & 34.03 & \em \small{+0.54} & 36.81 & \em \small{+0.61} & 2.39 \\
\bf {Our Method} & \bf{24.35*} & \bf{\em \small{+0.60}} & \bf{26.80**} & \bf{\em \small{+0.80}} & 1.97 & \bf{34.25**} & \bf{\em \small{+0.76}} & \bf{37.34*} & \bf{\em \small{+1.14}} & 1.57\\
\hline
\end{tabular}
}
\caption{BLEU scores on big data sets. *  and ** mean the improvements over  \textsc{Transformer} is statistically significant \cite{collins2005clause} ($\rho < 0.05$ and $\rho < 0.01$, respectively).} \label{tab-total-big}
\end{table*}

{\bf Our Method} Implemented based on Fairseq-py. The weight $\lambda$ in Equation \ref{eq-loss} for the small Chinese$\rightarrow$English data set is set to 0.25, and for other data sets is set to 0.5.
%True
%For the weight $\lambda$ in Equation \ref{eq-loss}, on the small Chinese$\rightarrow$English and English$\rightarrow$German data sets, $\lambda =0.5$, and on the English$\rightarrow$Romanian data set,  $\lambda =0.25$ and on the big Chinese$\rightarrow$English data set, $\lambda =0.3$.

%The same configuration for the Transformer-based systems. The number of multi heads is $8$, and the encoder and the decoder both have $6$ layers. The word embedding dimension and the hidden size $d_{model}$ are both $512$. Dropout is applied on embeddings, the output of the encode and the decoder all at the rate of $0.1$. batch-size 4096
All the Transformer-based systems have the same configuration as the base model described in \citet{vaswani2017attention} except that dropout rate is $0.3$.
%except that the dropout rate on the small Chinese$\rightarrow$English data set is 0.3 and that on other data sets is all 0.1.
The translation quality was evaluated with BLEU \cite{papinenibleu} with $n \mathcal{=} 4$ using the {\em SacreBLEU} tool~\cite{post-2018-call}\footnote{BLEU+case.mixed+numrefs.1+smooth.exp+tok.13a+ version.1.3.6}, where small data sets employ case-insensitive BLEU while big data sets use case-sensitive BLEU.

\subsection{Main Results} 

We compare our method with other methods that can make global planning, including the reinforcement-based method (\textsc{RL-NMT}),
the two-pass decoding method (\textsc{ABDNMT}), twin networks which match past and future information (\textsc{Twinnet}) and the NMT model with an evaluate module to evaluate fluency and faithfulness (\textsc{EVANMT}). In addition, we also explore learning mechanisms which can transfer knowledge from the seer decoder to the conventional decoder, including L$_2$ regularization (\textsc{Seer+L$_2$}),  adversarial learning (\textsc{Seer+AL}) and knowledge distillation (Our Method). 

We report results together with training time on the small and big data sets in Table \ref{tab-total-small} and Table \ref{tab-total-big}, respectively.\footnote{Please note that there is no comparability between our results and that of \citet{zhang2019future} because we used different validation and test sets.}  As for different methods, in the small data sets, ${\bf \textsc{RL-NMT}}$ can only get small improvements over Transformer which are in line with the results reported in \citet{wu2018study}, and  \textsc{ABDNMT} cannot get consistent improvements over Transformer with an obvious difference on the EN$\rightarrow$RO data set and a small difference on the CN$\rightarrow$EN data set.  
\textsc{Twinnet} can get comparable BLEU scores with our method on the small data sets but mostly negative difference on the big data sets. \textsc{EVANMT} can achieve consistent improvements and greater improvements on the 
EN$\rightarrow$DE data set.
For the learning mechanisms, knowledge distillation show consistent superiority over L$_2$ regularization and adversarial learning, which is remarkable especially on the big data sets. Adversarial learning can bring improvements over Transformer on all the data sets while L$_2$ regularization acts unstable on the big data sets. In summary, our method proved to be effective not only in the term of the architecture but also in the learning mechanism.

% table uuper bound

\begin{table*}[t!]
\centering
\scalebox{0.82}{
\begin{tabular}{l | c | c | c | c | c | c}
\hline
& \bf \textsc{MT03} & \bf \textsc{MT04} & \bf \textsc{MT05} &  \bf \textsc{MT06}  & \bf \textsc{MT08} & \bf \textsc{avg} \\
\hline
\hline
\bf \textsc{CD} \em{w/o} \textsc{CA}  & 6.24 & 6.68 & 6.70 & 6.77 & 4.49 & 6.12 \\
\bf \textsc{SD} \em{w/o} \textsc{CA}  & 16.39 & 16.70 & 16.64 & 17.21 & 11.97 & 15.78 \\
\hline
\bf \textsc{CD} \em{with} \textsc{CA}  & 29.45 & 25.03 & 30.14 & 32.07 & 23.39 & 28.02 \\
\bf \textsc{SD} \em{with} \textsc{CA} & 52.61 & 45.57 & 52.02 & 52.68 & 44.14 & 49.40 \\
\hline
\end{tabular}
}
\caption{BLEU scores of teacher forcing and seer forcing with and without cross-attention on NIST CN$\rightarrow$EN translation. \textsc{CD} and \textsc{SD} denote the conventional decoder and the seer decoder, respectively. \textsc{CA} represents cross-attention.} \label{tab-upper}
\end{table*}

\subsection{ The Superiority of the Seer Decoder} \label{sec-upper}

To use seer forcing to guide teacher forcing, it should be ensured that the seer decoder can outperform the conventional decoder. 
To verify this, we trained the two decoders together with the loss $\mathcal{L} = \mathcal{L}_t + \mathcal{L}_s$ without knowledge distillation. Then we evaluated their performance on the small Chinese-English translation task as follows. Both decoders are fed with ground truth words as context at test so that they can inference in the same way as at training, where the conventional decoder uses the past ground truth as context and the seer decoder employs  the past and future ground truth words as context in the past and future subdecoders. 

Besides translation performance, we also check the superiority of seer decoder in target language modeling. We do this by dropping out cross-attention so that the decoder can only generate translation based on target language model. In this way, the translation performance without cross-attention can demonstrate the ability of the two decoders in target language modeling.

%As the seer decoder is fed with past and future ground truth words, there is a possibility that it ignores the source information and degrades the translation task to a language modeling task. If so, the seer decoder cannot learn a good representation for translation. To further check this, we let the two decoders predict target words like language modeling without cross attention, then evaluated their BLEU scores.
 
We used the first reference of the test set as ground truth and calculated BLEU scores only with this reference. From the results in Table \ref{tab-upper}, we can see that whether with or without cross-attention the seer decoder can make super large improvements over the conventional decoder consistently on all the test sets. However, without cross-attention, the BLEU scores of both decoders decrease dramatically which means language model information is not enough for the translation task. 
Therefore, we can conclude the seer decoder acts much better  in target language modeling and cross-language projection and it is reasonable to use the seer decoder as the guider.

\iffalse  
\begin{table}[t!]
\centering
\scalebox{0.8}{
\begin{tabular}{l | l | c }
\hline
& \bf{Decoder} & \bf{AVG} \\
\hline
\hline
\multirow{2}{*}{\bf{\em with}} & \bf \textsc{Teacher}  & 28.02 \\
& \bf \textsc{Seer} & 49.40 \\
\hline
\multirow{2}{*}{\bf{\em without}} & \bf \textsc{Teacher}  &  6.12 \\
& \bf \textsc{Seer} & 15.78 \\
% MT02: 31.41 52.53
\hline
\end{tabular}
}
\caption{The upper-bound average BLEU scores of teacher forcing and seer forcing with and without cross-attention on cn$\rightarrow$en translation} \label{tab-upper}
\end{table}
\fi

\subsection{The Distillation of Future Information}

% table bag of words
\begin{table}[t!]
\centering
\resizebox{0.9\columnwidth}!{
\begin{tabular}{l | c | c | c }
\hline
& \bf {Accuracy} & \bf {Recall} & \bf {F1-Score} \\
\hline
\hline
\bf \textsc{Transformer}  & 47.23 & 40.91 & 43.84 \\
\bf {Our Method} & 52.24 & 42.10 & 46.63 \\
% MT02: 31.41 52.53
\hline
\end{tabular}
}
\caption{Comparison on the predicted bag of words between the conventional decoders} \label{tab-bows}
\end{table}

As the seer decoder achieves its superiority with the help of future target information, we hope that the conventional  decoder can learn future information from the seer decoder with knowledge distillation. To check this, we tested whether the hidden states of the conventional  decoder could derive more future ground truth words after knowledge distillation. The underlying belief is that the future ground information transferred from the seer decoder can help the conventional  decoder derive more future ground truth words.

%For the first experiment, we first trained the model with $\mathcal{L} = \mathcal{L}_t + \mathcal{L}_s$ (before ),  then added the knowledge distilation loss and trained it with Equation \ref{eq-loss} (after). We tested with the teacher  and seer decoders  before and after knowledge distillation on the human aligned data set from \citet{liu2015contrastive} which contains 900 Chinese-English sentence pairs. Here the two decoders were fed ground truth as in Section \ref{sec-upper}. For each target position, we selected as the alignment the link with the greatest probability, the quality of which was evaluated with Alignment Error Rate (AER) \cite{och2003minimum}. In the results shown in Table \ref{tab-aer}, the seer decoder has much lower AER than the teacher decoder which futher verifies it can learn better representation. After adversarial learning, the AER of the teacher decoder declines obviously while the AER of the seer decoder only increases slightly which indicates the teacher decoder approaches the seer decoder more and has learned better representations.

Assuming the hidden states generated by the conventional  decoder are $\bm{\mathrm{S}}_t=[\bm{\mathrm{s}}_{t_1}; ...; \bm{\mathrm{s}}_{t_I}]^T$, 
the future words for each target position $i$ can be predicted with the distribution
\begin{equation} 
\mathbf{P}_{wi} \sim \mathrm{softmax} (\mathbf{W}_w \mathbf{s}_{ti}) \label{eq-bow}
\end{equation} 
where $\mathbf{W}_w$ is the weight matrix. During training, we can get the bag of ground truth words for position i as  $\mathbf{y}_i^*=\{y_{i+1}^*, ..., y_I^*\}$ and train $\mathbf{W}_w$ with other parameters fixed by maximizing the likelihood of $\mathbf{y}_i^*$ as 
 \begin{equation}
\mathcal{L}_w = - \sum_{k=1}^K \sum_{i=1}^{I_k} \sum_{w \in \mathbf{y}_i^*} \log{p_{wi}(w)}
\end{equation}
where $K$ is the size of training sentences, $I_k$ is the length of the target sentence and $\log{p_{wi}(w)}$ is the probability of the word $w$ in Equation \ref{eq-bow}. 

At test, we select the top best $I_{b_i}$ words according to Equation \ref{eq-bow} as the bag of future words $\mathbf{b}_i$ for position $i$. As we cannot get the ground truth, the size of $\mathbf{b}_i$ is calculated approximately as 
$I_{b_i} = \max{\{2, (J-i) \times 2\}}$ where $J$ is the length of source sentence. As we do not know the target length during prediction, it may occur that $i$ is greater than  $J$ and calculating $I_{b_i}$ in this way can ensure $\mathbf{b}_i$ contains $2$ words at least.

We conducted experiments on Chinese-English translation and used MT02 as the test set only with the first reference as ground truth. We calculated the accuracy and recall by comparing each $\mathbf{b}_i$ against each $\mathbf{y}_i^*$.
The results in Table \ref{tab-bows} show the conventional  decoder in our method can achieve higher accuracy and recall compared to the decoder of Transformer. This means knowledge distillation does transfer future information from the seer decoder to the conventional  decoder.

\iffalse
\begin{table*}[t!]
\centering
\resizebox{1.45\columnwidth}!{
\begin{tabular}{l | c | c | c | c | c | c | c}
\hline
& \bf \textsc{MT03} & \bf \textsc{MT04} & \bf \textsc{MT05} &  \bf \textsc{MT06}  & \bf \textsc{MT08} & \bf \textsc{avg}  & \bf $\Delta$ \\
\hline
\hline
\bf {Our Method} & 48.12 & 48.85 & 48.25 & 47.25 & 38.71 & 46.24 &  \\
\hline
\quad \bf \textsc{-future}  & 47.29 & 48.20 & 47.33 & 46.36 & 37.74 & 45.38 &  \em \small{-0.86} \\
\quad \bf \textsc{-past}  & 47.91 & 48.04 & 47.01 & 46.59 & 37.53 & 45.42 & \em \small{-0.82} \\
\quad \bf \textsc{-kd} & 46.52 & 47.22 & 47.01 & 46.10 & 37.36 & 44.84 &  \em \small{-1.40} \\ 
\bf \textsc{Transformer}  & 46.54 & 46.95 & 46.39 & 45.39 & 36.75 & 44.40 & \em \small{-1.84} \\
\hline
\end{tabular}
}
\caption{Ablation study on NIST CN$\rightarrow$EN translation. \textsc{-future} : dropping the future subdecoder; \textsc{-past}: dropping the past subdecoder;  \textsc{-kd}: dropping knowledge distillation.} \label{tab-ablation}
\end{table*}
\fi

\begin{figure}[!t]
    \centering
    \includegraphics[scale=0.25]{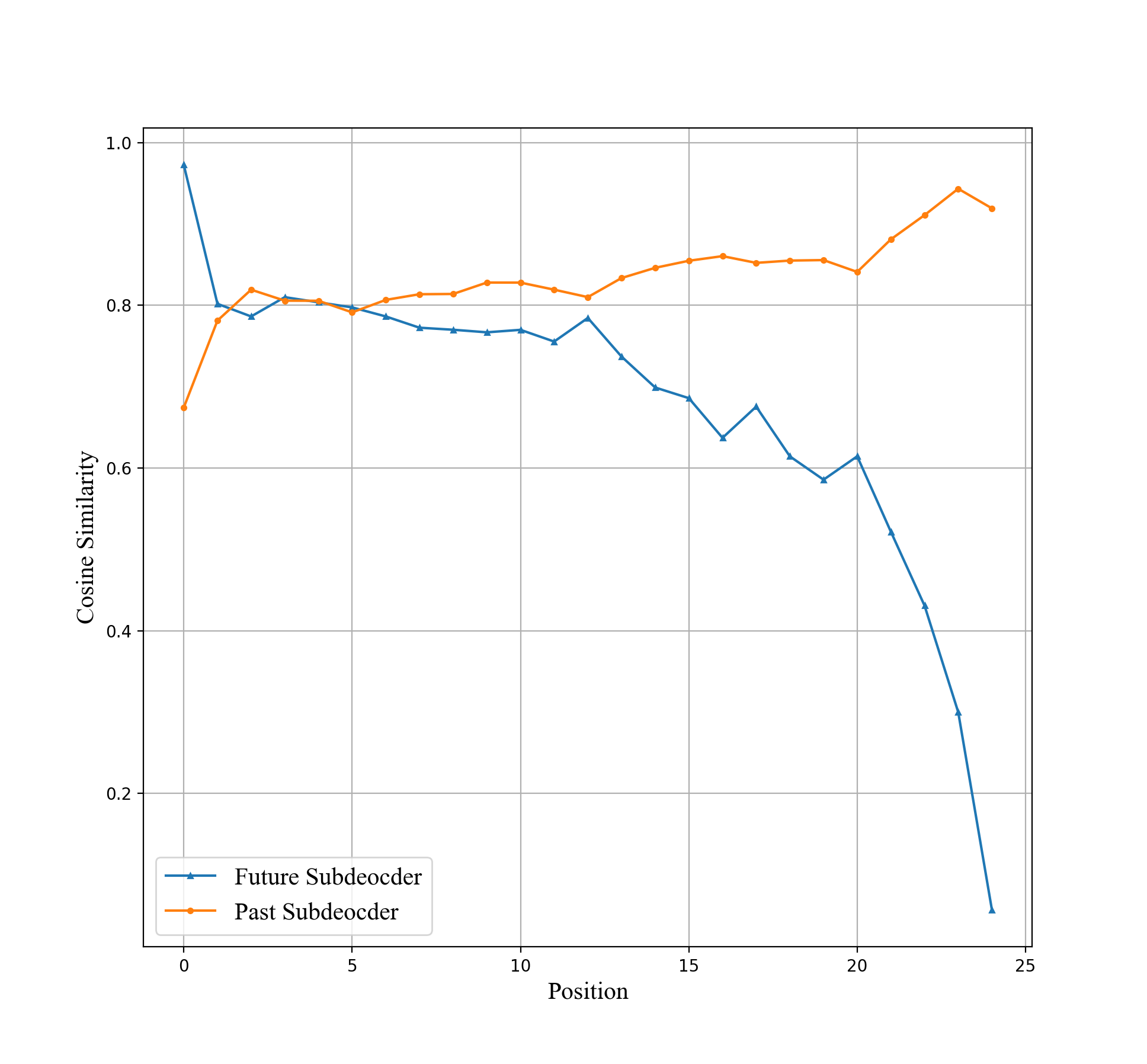}
    \caption{ The similarity of the past and future information to the fused information }
    \label{fig-ratio}
\end{figure}

\begin{table}[t!]
\centering
\resizebox{0.6\columnwidth}!{
\begin{tabular}{l | c | c}
\hline
& \bf \textsc{avg}  & \bf $\Delta$ \\
\hline
\hline
\bf {Our Method} & 46.24 &  \\
\hline
\quad \bf \textsc{-future}  &  45.38 &  \em \small{-0.86} \\
\quad \bf \textsc{-past}  & 45.42 & \em \small{-0.82} \\
\quad \bf \textsc{-kd} & 44.84 &  \em \small{-1.40} \\ 
\bf \textsc{Transformer}  & 44.40 & \em \small{-1.84} \\
\hline
\end{tabular}
}
\caption{Ablation study on NIST CN$\rightarrow$EN translation. \textsc{-future} : dropping the future subdecoder; \textsc{-past}: dropping the past subdecoder;  \textsc{-kd}: dropping knowledge distillation.} \label{tab-ablation}
\end{table}

\subsection{The Contribution of Subdecoders}

In the seer decoder of our method, the information from the past and future subdecoders is fused (as shown in Equation \ref{eq-fuse}) to get the final cross-attention. The intuition is that at the beginning stage, the past subdecoder contains less information than the future subdecoder, so the fused information should rely more on the future subdecoder. As the translation gets longer, the information embodied in the past subdecoder grows, and the fused information should depend more on the past subdecoder. To confirm this hypothesis, we calculate the cosine similarity of the vectors in $\mathbf{A}$ given in Equation \ref{eq-fuse} with the corresponding weighted vectors of $\bm{\mathrm{W}}_p\bm{\mathrm{H}}'_p$ and $\bm{\mathrm{W}}_f \bm{\mathrm{H}}''_f$.

We selected 205 sentences the length of which ranges $[15, 25]$, then calculated the cosine similarities word by word. Then the similarities at the same target position will be averaged and the chart over all the target positions is given in Figure \ref{fig-ratio}. The figure confirms our conjecture that at first, the fused information is highly related to the future information, and over time the similarity to past information increases gradually while the similarity to future information decreases faster.

\begin{figure}[!t]
    \centering
    \includegraphics[scale=0.35]{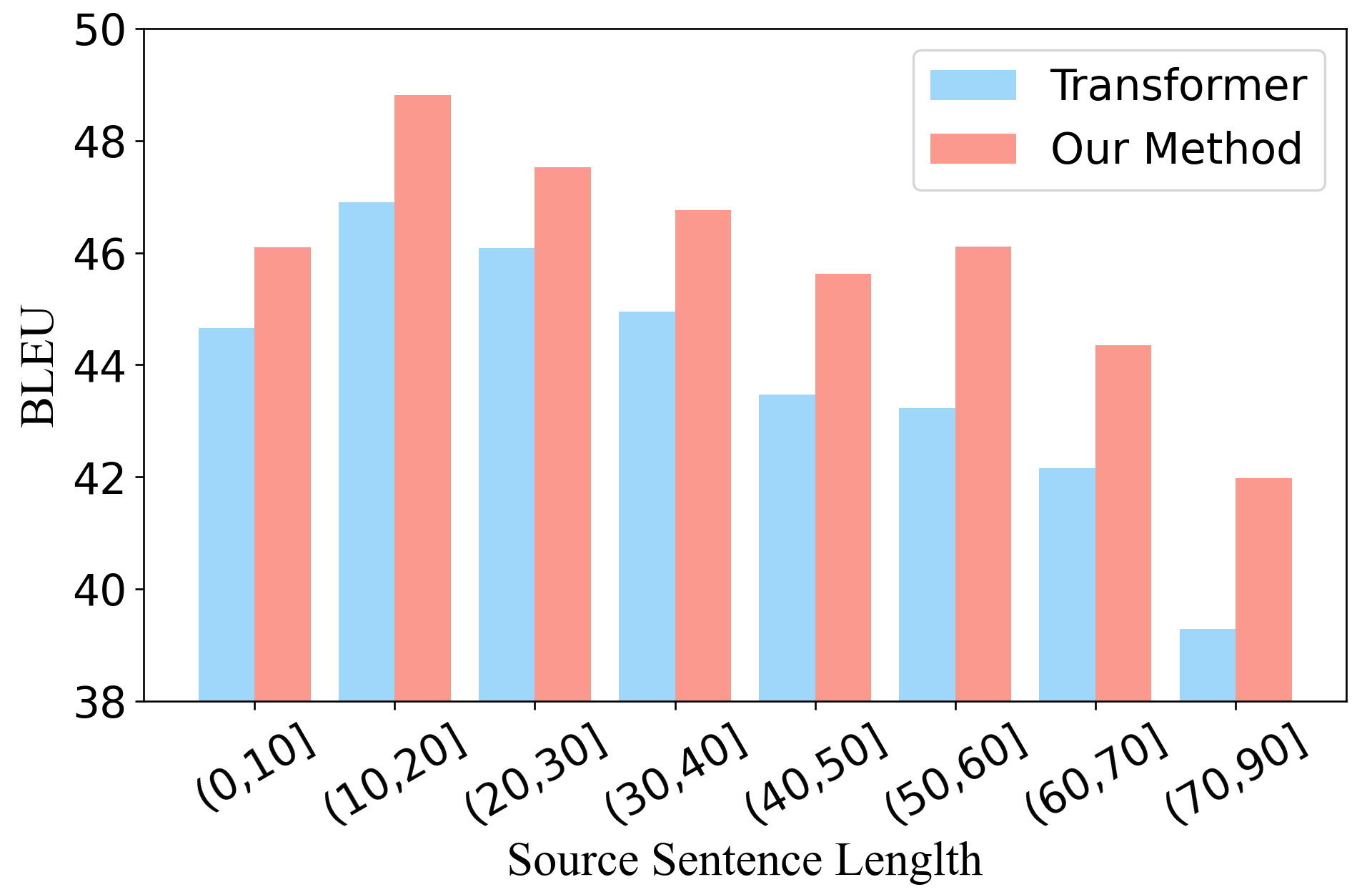}
    \caption{ The BLEU scores on sentence bins with different lengths.}
    \label{fig-len}
\end{figure}

\subsection{Ablation Study}

We have proven that in our method the past and future information collaborate to achieve better global planning. In this section, we will explore the influence of past and future information by separately deleting the {\em future} and {\em past} subdecoders from the seer decoder. In both cases, only the structure of the seer decoder changes and the whole model is trained with knowledge distillation in the same way. 
We also remove knowledge distillation loss in which case the seer and conventional  decoders only interact via the shared encoder and only optimize their own cross-entropy losses during training. 
The results are given in Table \ref{tab-ablation}.

When we exclude future or past information, the translation performance decreases dramatically at almost the same extent, but they still have an obvious gain compared to Transformer. This demonstrates that both the past and future information are necessary for global planning. It is interesting that the translation performance still rise without future subdecoder where there is no additional information fed compared to Transformer. The reason may be the conventional  and seer decoder can restrict each other to avoid bad behaviors. When knowledge distillation is 
dropped, the performance decline greatly which means only communicating via the encoder the conventional  and seer decoders is not enough. Hence we need to introduce knowledge distillation to reinforce the influence of the seer decoder to the conventional decoder.

\subsection{Performance with Sentence Length}

As the translation is generated word by word, the translation errors will be accumulated while the the translation grows, which will influence the later prediction. In our method, the conventional decoder can learn future information from the seer decoder and hence it should make better global planning for the whole sequence. From this, we deduce that our method performs better on long sentences than Transformer. 

We checked this on the NIST CN$\rightarrow$EN translation task and split the sentences in all the test sets  into 8 bins according to their length. Then we translated for each bin and tested the BLEU scores. The results in Figure \ref{fig-len} show that our method can achieve bigger improvements on longer sentences, especially in the last three bins. 

%we compared our system with related methods. As our method and the method under the reinforcement framework both inference with the conventional decoder (the teacher decoder), we report the comparison result with this method. We also check the effect of adopting a discriminator rather than the L2 regularization. Besides, we also list the results of our implemented \textsc{rnnsearch} and \textsc{abdnmt} based on \textsc{rnnsearch}. 

%The results in Table \ref{tab-cnen} and Table \ref{tab-dero} shown that the usage of the seer decoder can indeed guide the teacher decoder to achieve better performance even with the L2 regularization and the discriminator shows its superiority over the L2 regularization.

\section{Conclusion}

In order to help the NMT model to make good global planning at inference, we propose to introduce a seer decoder which embodies future ground truth to guide the behaviors of the conventional  decoder. To this end, we employ the method of knowledge distillation to transfer future information from the seer decoder to the conventional  decoder. At test, the conventional decoder can perform translation on its own as if it knew some future information.
The experiments indicate our method can outperform strong baselines significantly on four data sets. We are also the first to explore learning mechanisms of knowledge distillation, adversarial learning and L$_2$ regularization and knowledge distillation has proven to be the most effective one.

\section*{Acknowledgement}

This paper was supported by National Key R\&D Program of China (NO. 2017YFE0192900). Thank Wanying Xie for running the experiments of \textsc{EVANMT}. 
We thank all the anonymous reviewers for the insightful and valuable comments.

\bibliographystyle{acl_natbib}
\bibliography{acl2021}

\end{document}